\documentclass[10pt,twocolumn,letterpaper]{article}

\usepackage{cvpr}
\usepackage{times}
\usepackage{epsfig}
\usepackage{graphicx}
\usepackage{amsmath}
\usepackage{amssymb}
\usepackage{microtype}
\usepackage[american]{babel}
\usepackage{bm}
\usepackage[caption=false]{subfig}
\usepackage{float}
\usepackage{multirow}
\usepackage{epstopdf}

\usepackage[pagebackref=true,breaklinks=true,letterpaper=true,colorlinks,bookmarks=false]{hyperref}

\cvprfinalcopy 

\def\cvprPaperID{3140} 

\ifcvprfinal\fi
\begin{document}

\title{Multi-Label Image Recognition with Graph Convolutional Networks\thanks{Z.-M. Chen's contribution was made when he was an intern in Megvii Research Nanjing. X.-S. Wei and Y. Guo are the corresponding authors. Yanwen Guo is also with the Science and Technology on Information Systems Engineering Laboratory, China, and the 28th Research Institute of China Electronics Technology Group Corporation, Nanjing 210007, China. This research was supported by National Key R\&D Program of China (No. 2017YFA0700800), the National Natural Science Foundation of China under Grants 61772257 and 61672279.
}}

\author{Zhao-Min Chen$^{1,2}$ \qquad Xiu-Shen Wei$^{2}$ \qquad Peng Wang$^{3}$ \qquad Yanwen Guo$^{1}$\\
\noindent $^1$National Key Laboratory for Novel Software Technology, Nanjing University, China\\
$^2$Megvii Research Nanjing, Megvii Technology, China\\
$^3$School of Computer Science, The University of Adelaide, Australia\\
{\tt\small \{chenzhaomin123, weixs.gm\}@gmail.com, peng.wang@adelaide.edu.au, ywguo@nju.edu.cn}\\}

\maketitle
\thispagestyle{empty}

\begin{abstract}
The task of multi-label image recognition is to predict a set of object labels that present in an image. As objects normally co-occur in an image, it is desirable to model the label dependencies to improve the recognition performance. To capture and explore such important dependencies, we propose a multi-label classification model based on Graph Convolutional Network (GCN). The model builds a directed graph over the object labels, where each node (label) is represented by word embeddings of a label, and GCN is learned to map this label graph into a set of inter-dependent object classifiers. These classifiers are applied to the image descriptors extracted by another sub-net, enabling the whole network to be end-to-end trainable. Furthermore, we propose a novel re-weighted scheme to create an effective label correlation matrix to guide information propagation among the nodes in GCN. Experiments on two multi-label image recognition datasets show that our approach obviously outperforms other existing state-of-the-art methods. In addition, visualization analyses reveal that the classifiers learned by our model maintain meaningful semantic topology.
\end{abstract}

\begin{figure}
	\centering
	\includegraphics[width=0.8\columnwidth]{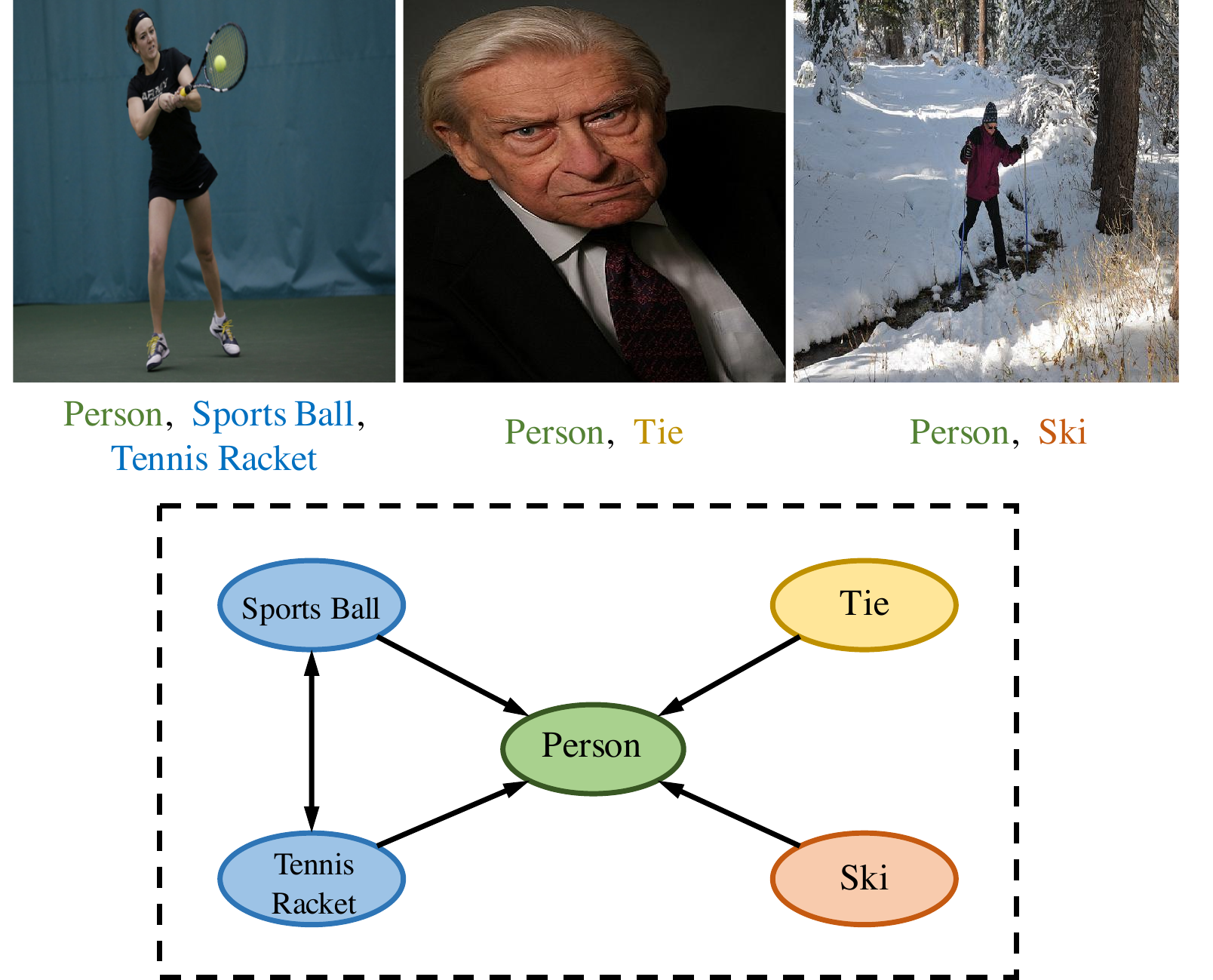}
	\caption{We build a directed graph over the object labels to model label dependencies in multi-label image recognition. In this figure, ``$\rm Label_{A} \rightarrow Label_{B}$'', means when $\rm Label_{A}$ appears, $\rm Label_{B}$ is likely to appear, but the reverse may not be true.}
	\label{fig:multi}
\end{figure}


\section{Introduction}\label{sec:intro}

Multi-label image recognition is a fundamental and practical task in Computer Vision, where the aim is to predict a set of objects present in an image. It can be applied to many fields such as medical diagnosis recognition~\cite{chest}, human attribute recognition~\cite{human} and retail checkout recognition~\cite{retail,rpc}. Comparing to multi-class image classification~\cite{7812753}, the multi-label task is more challenging due to the combinatorial nature of the output space. As the objects normally co-occur in the physical world, a key for multi-label image recognition is to model the label dependencies, as shown in Fig.~\ref{fig:multi}.

A Na\"ive way to address the multi-label recognition problem is to treat the objects in isolation and convert the multi-label problem into a set of binary classification problems to predict whether each object of interest presents or not. Benefited from the great success of  single-label image classification achieved by deep Convolutional Neural Networks (CNNs)~\cite{resnet,verydeep,inception,densenet}, the performance of the binary solutions has been greatly improved. However, these methods are essentially limited by ignoring the complex topology structure between objects. This stimulates research for approaches to capture and explore the label correlations in various ways. Some approaches, based on probabilistic graph model~\cite{tree,conditional} or Recurrent Neural Networks (RNNs)~\cite{cnn-rnn}, are proposed to explicitly model label dependencies. While the former formulates the multi-label classification problem as a structural inference problem which may suffer from a scalability issue due to high computational complexity, the latter predicts the labels in a sequential fashion, based on some orders either pre-defined or learned. Another line of works implicitly model the label correlations via attention mechanisms~\cite{srn,rnn_attention}. They consider the relations between attended regions of an image, which can be viewed as \emph{local correlations}, but still ignore the \emph{global correlations} between labels which require to be inferred from knowledge beyond a single image.   
 
In this paper, we propose a novel GCN based model (\emph{aka} ML-GCN) to capture the label correlations for multi-label image recognition, which properties with scalability and flexibility impossible for competing approaches. Instead of treating object classifiers as a set of independent parameter vectors to be learned, we propose to learn inter-dependent object classifiers from prior label representations, \eg, word embeddings, via a GCN based mapping function. In the following, the generated classifiers are applied to image representations generated by another sub-net to enable end-to-end training. As the embedding-to-classifier mapping parameters are shared across all classes (\ie, image labels), the gradients from all classifiers impact the GCN based classifier generation function. This implicitly models the label correlations. Furthermore, to explicitly model the label dependencies for classifier learning, we design an effective label correlation matrix to guide the information propagation among nodes in GCN. Specifically, we propose a re-weighted scheme to balance the weights between a node and its neighborhood for node feature update, which effectively alleviates overfitting and over-smoothing. Experiments on two multi-label image recognition datasets show that our approach obviously outperforms existing state-of-the-art methods. In addition, visualization analyses reveal that the classifiers learned by our model maintain meaningful semantic structures.


The main contributions of this paper are as follows:
\vspace{-5pt}
 \begin{itemize}
\setlength{\itemsep}{-2pt}
 	\item We propose a novel end-to-end trainable multi-label image recognition framework, which employs GCN to map label representations, \eg, word embeddings, to inter-dependent object classifiers.
 	\item We conduct in-depth studies on the design of correlation matrix for GCN and propose an effective re-weighted scheme to simultaneously alleviate the over-fitting and over-smoothing problems.
 	\item We evaluate our method on two benchmark multi-label image recognition datasets, and our proposed method consistently achieves superior performance over previous competing approaches.
 \end{itemize}
 
\section{Related Work}

The performance of image classification has recently witnessed a rapid progress due to the establishment of large-scale hand-labeled datasets such as ImageNet~\cite{imagenet}, MS-COCO~\cite{coco} and PASCAL VOC~\cite{voc}, and the fast development of deep convolutional networks~\cite{resnet,senet,shufflenet,xception,resnext}. Many efforts have been dedicated to extending deep convolutional networks for multi-label image recognition.

A straightforward way for multi-label recognition is to train independent binary classifiers for each class/label. However, this method does not consider the relationship among labels, and the number of predicted labels will grow exponentially as the number of categories increase. For instance, if a dataset contains 20 labels, then the number of predicted label combination could be more than 1 million (\emph{i.e.}, $2^{20}$). Besides, this baseline method is essentially limited by ignoring the topology structure among objects, which can be an important regularizer for the co-occurrence patterns of objects. For example, some combinations of labels are almost impossible to appear in the physical world.


In order to regularize the prediction space, many researchers attempted to capture label dependencies. Gong \emph{et al.}~\cite{warp} used a ranking-based learning strategy to train deep convolutional neural networks for multi-label image recognition and found that the weighted approximated-ranking loss worked best. Additionally, Wang \emph{et al.}~\cite{cnn-rnn} utilized recurrent neural networks (RNNs) to transform labels into embedded label vectors, so that the correlation between labels can be employed. Furthermore, attention mechanisms were also widely applied to discover the label correlation in the multi-label recognition task. In~\cite{srn}, Zhu \emph{et al.} proposed a spatial regularization network to capture both semantic and spatial relations of these multiple labels based on weighted attention maps. Wang \emph{et al.}~\cite{rnn_attention} introduced a spatial transformer layer and long short-term memory (LSTM) units to capture the label correlation.

Compared with the aforementioned structure learning methods, the \emph{graph} was proven to be more effective in modeling label correlation. Li \emph{et al.}~\cite{tree} created a tree-structured graph in the label space by using the maximum spanning tree algorithm. Li \emph{et al.}~\cite{conditional} produced image-dependent conditional label structures base on the graphical Lasso framework. Lee \emph{et al.}~\cite{ml-zsl} incorporated knowledge graphs for describing the relationships between multiple labels. In this paper, we leverage the graph structure to capture and explore the label correlation dependency. Specifically, based on the graph, we utilize GCN to propagate information between multiple labels and consequently learn inter-dependent classifiers for each of image labels. These classifiers absorb information from the label graph, which are further applied to the global image representation for the final multi-label prediction. It is a more explicit way for evaluating label co-occurrence. Experimental results validate our proposed approach is effective and our model can be trained in an end-to-end manner. 

\begin{figure*}
	\centering
	\includegraphics[width=2.0\columnwidth]{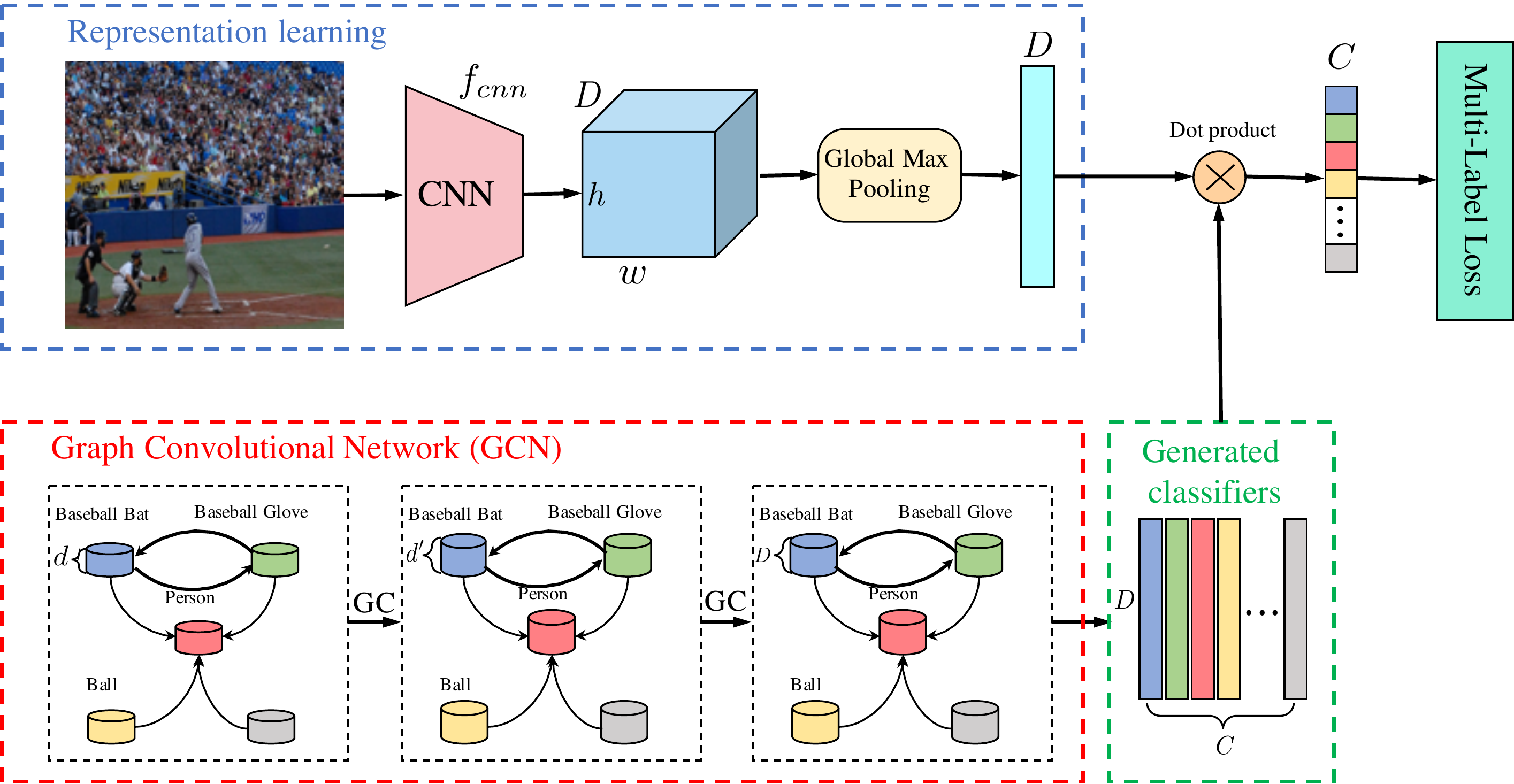}
	\vspace{0.5em}
	\caption{Overall framework of our ML-GCN model for multi-label image recognition.
The object labels are represented by {word embeddings} $\bm{Z} \in \mathbb{R}^{C \times d}$ ($C$ is the number of categories and $d$ is the dimensionality of word-embedding vector). A directed graph is built over these label representations, where each node denotes a label. Stacked GCNs are learned over the label graph to map these label representations into a set of inter-dependent object classifiers, \ie, $\bm{W\in\mathbb{R}^{C\times{D}}}$, which are applied to the image representation extracted from the input image via a convolutional network for multi-label image recognition.}

	\vspace{-0.5em}
	\label{fig:model}
\end{figure*}

\section{Approach}

In this part, we elaborate on our ML-GCN model for multi-label image recognition. Firstly, we introduce the motivation for our method. Then, we introduce some preliminary knowledge of GCN, which is followed by the detailed illustration of the proposed ML-GCN model and the re-weighted scheme for correlation matrix construction.

\subsection{Motivations}\label{sec:motivation}

How to effectively capture the correlations between object labels and explore these label correlations to improve the classification performance are both important for multi-label image recognition. In this paper, we use a graph to model the inter dependencies between labels, which is a flexible way to capture the topological structure in the label space. Specifically, we represent each node (label) of the graph as word embeddings of the label, and propose to use GCN to directly map these label embeddings into a set of inter-dependent classifiers, which can be directly applied to an image feature for classification. Two factors motivated the design of our GCN based model. Firstly, as the embedding-to-classifier mapping parameters are shared across all classes, the learned classifiers can retain the weak semantic structures in the word embedding space, where semantic related concepts are close to each other. Meanwhile, the gradients of all classifiers can impact the classifier generation function, which implicitly models the label dependencies. Secondly,  we design a novel label correlation matrix based on their co-occurrence patterns to explicitly model the label dependencies by GCN, with which the update of node features will absorb information from correlated nodes (labels). 


\subsection{Graph Convolutional Network Recap}

Graph Convolutional Network (GCN) was introduced in~\cite{gcn} to perform semi-supervised classification. The essential idea is to update the node representations by propagating information between nodes.

Unlike standard convolutions that operate on local Euclidean structures in an image, the goal of GCN is to learn a function $f(\cdot,\cdot)$ on a graph $\mathcal{G}$, which takes feature descriptions $\bm{H}^{l}\in\mathbb{R}^{n\times{d}}$ and the corresponding correlation matrix $\bm{A} \in \mathbb{R}^{n \times n}$ as inputs (where $n$ denotes the number of nodes and $d$ indicates the dimensionality of node features), 
and updates the node features as $\bm{H}^{l+1} \in \mathbb{R}^{n \times d'}$. Every GCN layer can be written as a non-linear function by
\begin{equation}
	\bm{H}^{l+1} = f(\bm{H}^{l}, \bm{A}).
\end{equation}
After employing the convolutional operation of~\cite{gcn},  $f(\cdot,\cdot)$ can be represented as
\begin{equation}
\label{eq:gcn}
	\bm{H}^{l+1} = h(\bm{\widehat{A}} \bm{H}^{l} \bm{W}^{l}),
\end{equation}
where $\bm{W}^{l} \in \mathbb{R}^{d \times d'}$ is a transformation matrix to be learned and $\bm{\widehat{A}}\in\mathbb{R}^{n\times{n}}$ is the normalized version of correlation matrix $\bm{A}$, and $h(\cdot)$ denotes a non-linear operation, which is acted by LeakyReLU~\cite{leakyrelu} in our experiments. Thus, we can learn and model the complex inter-relationships of the nodes by stacking multiple GCN layers. For more details, we refer interested readers to~\cite{gcn}.

\subsection{GCN for Multi-label Recognition}

Our ML-GCN is built upon GCN. GCN was proposed for semi-supervised classification, where the node-level output is the prediction score of each node. Different from that, we design the final output of each GCN node to be the classifier of the corresponding label in our task. In addition, the graph structure (\ie, the correlation matrix) is normally pre-defined in other tasks, which, however, is not provided in the multi-label image recognition task. Thus, we need to construct the correlation matrix from scratch. The overall framework of our approach is shown in Fig.~\ref{fig:model}, which is composed of two main modules, \ie, the image representation learning and GCN based classifier learning modules.


\paragraph{Image representation learning} We can use any CNN base models to learn the features of an image. In our experiments, following~\cite{srn,order,ml-zsl,multi_evidence}, we use ResNet-101~\cite{resnet} as the base model in experiments. Thus, if an input image $\bm{I}$ is with the $448 \times 448$ resolution, we can obtain $2048 \times 14 \times 14$ feature maps from the ``\texttt{conv5\_x}'' layer. Then, we employ global max-pooling to obtain the image-level feature $\bm{x}$:
\begin{equation}
	\bm{x} = f_{\rm GMP}(f_{\rm cnn} (\bm{I}; \theta_{\rm cnn})) \in \mathbb{R}^{D},
\end{equation}
where $\theta_{\rm cnn}$ indicates model parameters and $D=2048$.

\paragraph{GCN based classifier learning} We learn inter-dependent object classifiers, \ie, $\bm{W} = \{\bm{w}_{i}\}_{i=1}^{C}$, from label representations via a GCN based mapping function, where $C$ denotes the number of categories. We use stacked GCNs where each GCN layer \textit{l} takes the node representations from previous layer ($\bm{H}^{l}$) as inputs and outputs new node representations, \ie, $\bm{H}^{l+1}$. For the first layer, the input is the $\bm{Z} \in \mathbb{R}^{C \times d}$ matrix, where $d$ is the dimensionality of the label-level word embedding. For the last layer, the output is $\bm{W} \in \mathbb{R}^{C\times{D}}$ with $D$ denoting the dimensionality of the image representation. By applying the learned classifiers to image representations, we can obtain the predicted scores as

\begin{equation}
	\hat{\bm{y}} = \bm{W} \bm{x}.
\end{equation}

We assume that the ground truth label of an image is $\bm{y}\in\mathbb{R}^{C}$, where $y^i = \{0,1\}$ denotes whether label $i$ appears in the image or not. 
The whole network is trained using the traditional multi-label classification loss as follows
\begin{equation}
	\mathcal{L}= \sum_{c=1}^{C}y^{c}\log(\sigma(\hat{y}^{c})) + (1-y^{c})\log(1-\sigma(\hat{y}^{c})),
\end{equation}
where $\sigma(\cdot)$ is the sigmoid function.

\subsection{Correlation Matrix of ML-GCN}\label{sec:cormat}

GCN works by propagating information between nodes based on the correlation matrix. Thus, how to build the correlation matrix $\bm{A}$ is a crucial problem for GCN. In most applications, the correlation matrix is pre-defined, which, however, is not provided in any standard multi-label image recognition datasets. In this paper, we build this correlation matrix through a data-driven way. That is, we define the correlation between labels via mining their co-occurrence patterns within the dataset. 

\begin{figure}[t!]
	\centering
	\includegraphics[width=0.95\columnwidth]{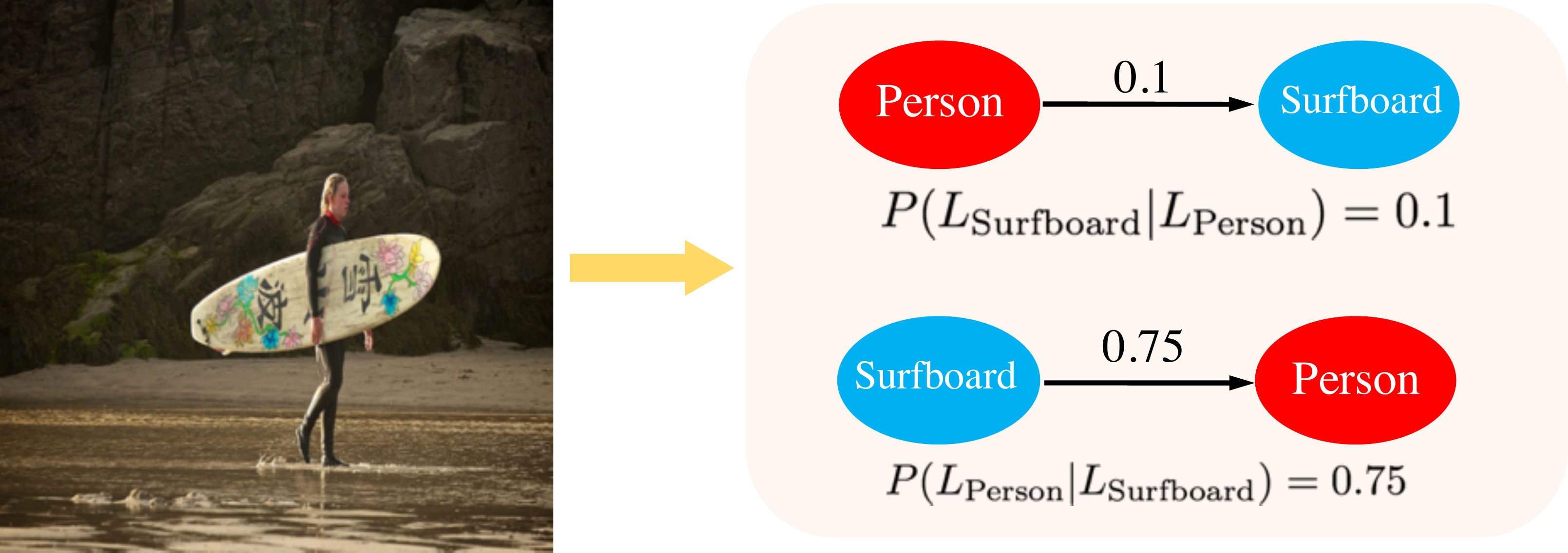}
	\caption{Illustration of conditional probability between two labels. As usual, when ``\texttt{surfboard}'' appears in the image, ``\texttt{person}'' will also occur with a high probability. However, in the condition of ``\texttt{person}'' appearing, ``\texttt{surfboard}'' will not necessarily occur.}
	\label{fig:condition}
\end{figure}


We model the label correlation dependency in the form of conditional probability, \ie, $P(L_{j}|L_{i})$ which denotes the probability of occurrence of label $L_{j}$ when label $L_{i}$ appears. As shown in Fig.~\ref{fig:condition}, $P(L_{j}|L_{i})$ is not equal to $P(L_{i}|L_{j})$. Thus, the correlation matrix is asymmetrical. 

To construct the correlation matrix, firstly, we count the occurrence of label pairs in the training set and get the matrix $\bm{M}\in\mathbb{R}^{C\times{C}}$. Concretely, $C$ is the number of categories, and $M_{ij}$ denotes the concurring times of $L_{i}$ and $L_{j}$. Then, by using this label co-occurrence matrix, we can get the conditional probability matrix by 

\begin{equation}
	\bm{P}_{i} = \bm{M}_{i} / N_{i},
\end{equation}
where $N_{i}$ denotes the occurrence times of $L_i$ in the training set, and $P_{ij} = P(L_{j}|L_{i})$ means the probability of label $L_j$ when label $L_i$ appears.

However, the simple correlation above may suffer from two drawbacks. Firstly, the co-occurrence patterns between a label and the other labels may exhibit a long-tail distribution, where some rare co-occurrences may be noise. Secondly, the absolute number of co-occurrences from training and test may not be completely consistent. A correlation matrix overfitted to the training set can hurt the generalization capacity. Thus, we propose to binarize the correlation $\bm{P}$. Specifically, we use the threshold $\tau$ to filter noisy edges, and the operation can be written as
\begin{equation}
\label{eq:binary}
	\bm{A}_{ij} =
\begin{cases} 
0,  & \mbox{if }\bm{P}_{ij} < \tau \\
1, & \mbox{if }\bm{P}_{ij} \geq \tau
\end{cases}\,,
\end{equation}
where $\bm{A}$ is the binary correlation matrix.

\paragraph{Over-smoothing problem} From Eq.~(\ref{eq:gcn}), we can conclude that after GCN, the feature of a node will be the weighted sum of its own feature and the adjacent nodes' features. Then, a direct problem for the binary correlation matrix is that it can result in over-smoothing. That is, the node features may be over-smoothed such that nodes from different clusters (\eg, kitchen related \vs living room related) may become indistinguishable~\cite{oversmooth}. To alleviate this problem, we propose the following re-weighted scheme,
\begin{equation}
\label{eq:reweight}
	\bm{A'}_{ij} = 
\begin{cases} 
p / \sum_{\substack{j=1 \\ i \neq j}}^{C} \bm{A}_{ij},  & \mbox{if }i \neq j \\
1 - p, & \mbox{if }i = j
\end{cases}\,,
\end{equation}
where $\bm{A'}$ is the re-weighted correlation matrix, and $p$ determines the weights assigned to a node itself and other correlated nodes. By doing this, when updating the node feature, we will have a fixed weight for the node itself and the weights for correlated nodes will be determined by the neighborhood distribution. When $p\to 1$, the feature of a node itself will not be considered. While, on the other hand, when $p\to 0$, neighboring information tends to be ignored.

\section{Experiments}

In this section, we first describe the evaluation metrics and implementation details. Then, we report the empirical results on two benchmark multi-label image recognition datasets, \ie, {MS-COCO}~\cite{coco} and {VOC}~2007~\cite{voc}. Finally, visualization analyses are presented.

\subsection{Evaluation Metrics}

Following conventional settings~\cite{cnn-rnn,multi_evidence,srn}, we report the average per-class precision (CP), recall (CR), F1 (CF1) and the average overall precision (OP), recall (OR), F1 (OF1) for performance evaluation. For each image, the labels are predicted as positive if the confidences of them are greater than 0.5. For fair comparisons, we also report the results of top-3 labels, cf.~\cite{srn,multi_evidence}. In addition, we also compute and report the mean average precision (mAP). Generally, average overall F1 (\textbf{OF1}), average per-class F1 (\textbf{CF1}) and \textbf{mAP} are relatively more important for performance evaluation.

\subsection{Implementation Details}

Without otherwise stated, our ML-GCN consists of two GCN layers with output dimensionality of $1024$ and $2048$, respectively.
For label representations, we adopt $300$-dim GloVe~\cite{glove} trained on the Wikipedia dataset. For the categories whose names contain multiple words, we obtain the label representation as average of embeddings for all words. For the correlation matrix, without otherwise stated, we set $\tau$ in Eq.~(\ref{eq:binary}) to be 0.4 and $p$ in Eq.~(\ref{eq:reweight}) to be 0.2. In the image representation learning branch, we adopt LeakyReLU~\cite{leakyrelu} with the negative slope of 0.2 as the non-linear activation function, which leads to faster convergence in experiments. We adopt ResNet-101~\cite{resnet} as the feature extraction backbone, which is
pre-trained on {ImageNet}~\cite{imagenet}.
During training, the input images are random cropped and resized into $448 \times 448$ with random horizontal flips for data augmentation. For network optimization, SGD is used as the optimizer. The momentum is set to be 0.9. Weight decay is $10^{-4}$. The initial learning rate is 0.01, which decays by a factor of 10 for every 40 epochs and the network is trained for 100 epochs in total. We implement the network based on PyTorch.


\subsection{Experimental Results}

In this part, we first present our comparisons with state-of-the-arts on MS-COCO and VOC 2007, respectively. Then, we conduct ablation studies to evaluate the key aspects of the proposed approach.

\subsubsection{Comparisons with State-of-the-Arts}

\paragraph{Results on MS-COCO}

Microsoft COCO~\cite{coco} is a widely used benchmark for multi-label image recognition. It contains 82,081 images as the training set and 40,504 images as the validation set. The objects are categorized into 80 classes with about 2.9 object labels per image. Since the ground-truth labels of the test set are not available, we evaluate the performance of all the methods on the validation set. The number of labels of different images also varies considerably, which makes MS-COCO more challenging.  

Quantitative results are reported in Table~\ref{table:coco}. We compare with state-of-the-art methods, including CNN-RNN~\cite{cnn-rnn}, RNN-Attention~\cite{rnn_attention}, Order-Free RNN~\cite{order}, ML-ZSL~\cite{ml-zsl}, SRN~\cite{srn}, Multi-Evidence~\cite{multi_evidence}, etc. For the proposed ML-GCN, we report the results based on the binary correlation matrix (``ML-GCN (Binary)'')  and the re-weighted correlation matrix (``ML-GCN (Re-weighted)''), respectively. It is obvious to see that our ML-GCN method based on the binary correlation matrix obtains worse classification performance, which may be largely due to the over-smoothing problem discussed in Sec.~\ref{sec:cormat}. The proposed re-weighted scheme can alleviate the over-smoothing issue and consequently obtains superior performance. Comparing with state-of-the-art methods, our approach with the proposed re-weighted scheme consistently performs better under almost all metrics, which shows the effectiveness of our proposed ML-GCN as well as its corresponding re-weighted scheme.


\begin{table*}[t]
\centering
\footnotesize
\caption{Comparisons with state-of-the-art methods on the MS-COCO dataset. The performance of the proposed ML-GCN  based on two types of correlation matrices are reported.``Binary'' denotes that we use the binary correlation matrix, cf.~Eq.~(\ref{eq:binary}). ``Re-weighted'' means the correlation matrix generated by the proposed re-weighted scheme is used, cf. Eq.~(\ref{eq:reweight}).}
\vspace{0.19cm}
\begin{tabular}{|c||c|c|c|c|c|c|c||c|c|c|c|c|c|}
\hline
\multirow{2}{*}{Methods} & \multicolumn{7}{c||}{{All}} & \multicolumn{6}{c|}{{Top-3}} \\
\cline{2-14} & mAP & CP & CR & CF1 & OP & OR & OF1 & CP & CR & CF1 & OP & OR & OF1 \\
\hline
\hline
CNN-RNN~\cite{cnn-rnn} & 61.2 & -- & -- & -- & -- & -- & -- & 66.0 & 55.6 & 60.4 & 69.2 & 66.4 & 67.8\\
RNN-Attention~\cite{rnn_attention} & -- & -- & -- & -- & -- & -- & -- & 79.1 & 58.7 & 67.4 & 84.0 & 63.0 & 72.0\\
Order-Free RNN~\cite{order} & -- & -- & -- & -- & -- & -- & -- & 71.6 & 54.8 & 62.1 & 74.2 & 62.2 & 67.7\\
ML-ZSL~\cite{ml-zsl} & -- & -- & -- & -- & -- & -- & -- & 74.1 & \textbf{64.5} & 69.0 & -- & -- & --\\
SRN~\cite{srn} & 77.1 & {81.6} & 65.4 & 71.2 & 82.7 & 69.9 & 75.8 & 85.2 & 58.8 & 67.4 & 87.4 & 62.5 & 72.9\\
ResNet-101~\cite{resnet} & 77.3 & 80.2 & 66.7 & 72.8 & 83.9 & 70.8 & 76.8 & 84.1 & 59.4 & 69.7 & 89.1 & 62.8 &73.6\\
Multi-Evidence~\cite{multi_evidence} & -- & 80.4 & 70.2 & 74.9 & 85.2 & 72.5 & 78.4 & 84.5 & 62.2 & 70.6 & {89.1} & 64.3 & 74.7\\
\hline
\hline
ML-GCN (Binary) & 80.3 & 81.1 & 70.1 & 75.2 & 83.8 & 74.2 & 78.7 & 84.9 & 61.3 & 71.2 & 88.8 & 65.2 & 75.2\\
ML-GCN (Re-weighted)  & \textbf{83.0} & \textbf{85.1} & \textbf{72.0} & \textbf{78.0} & \textbf{85.8} & \textbf{75.4} & \textbf{80.3} & \textbf{89.2} & 64.1 & \textbf{74.6} & \textbf{90.5} & \textbf{66.5} & \textbf{76.7}\\
\hline
\end{tabular}
\label{table:coco}
\end{table*}

\begin{table*}[t]
\centering
\footnotesize
\setlength{\tabcolsep}{0.64mm}
\caption{Comparisons of AP and mAP with state-of-the-art methods on the {VOC 2007} dataset. The meanings of ``Binary'' and ``Re-weighted'' are the same as Table~\ref{table:coco}.}
\vspace{0.18cm}
\begin{tabular}{|c||c|c|c|c|c|c|c|c|c|c|c|c|c|c|c|c|c|c|c|c||c|}
\hline
Methods & \textit{aero} & \textit{bike} & \textit{bird} & \textit{boat} & \textit{bottle} & \textit{bus} & \textit{car} & \textit{cat} & \textit{chair} & \textit{cow} & \textit{table} & \textit{dog} & \textit{horse} & \textit{motor} & \textit{person} & \textit{plant} & \textit{sheep} & \textit{sofa} & \textit{train} & \textit{tv} & mAP\\
\hline
\hline
CNN-RNN~\cite{cnn-rnn} & 96.7 & 83.1 & 94.2 & 92.8 & 61.2 & 82.1 & 89.1 & 94.2 & 64.2 & 83.6 & 70.0 & 92.4 & 91.7 & 84.2 & 93.7 & 59.8 & 93.2 & 75.3 & \textbf{99.7} & 78.6 & 84.0\\
RLSD~\cite{rlsd} & 96.4 & 92.7 & 93.8 & 94.1 & 71.2 & 92.5 & 94.2 & 95.7 & 74.3 & 90.0 & 74.2 & 95.4 & 96.2 & 92.1 & 97.9 & 66.9 & 93.5 & 73.7 & 97.5 & 87.6 & 88.5\\
VeryDeep~\cite{verydeep} & 98.9 & 95.0 & 96.8 & 95.4 & 69.7 & 90.4 & 93.5 & 96.0 & 74.2 & 86.6 & \textbf{87.8} & 96.0 & 96.3 & 93.1 & 97.2 & 70.0 & 92.1 & 80.3 & 98.1 & 87.0 & 89.7\\
ResNet-101~\cite{resnet} & 99.5 & 97.7 & 97.8 & 96.4 & 65.7 & 91.8 & 96.1 & 97.6 & 74.2 & 80.9 & 85.0 & \textbf{98.4} & 96.5 & 95.9 & 98.4 & 70.1 & 88.3 & 80.2 & 98.9 & 89.2 & 89.9\\
FeV+LV~\cite{fev} & 97.9 & 97.0 & 96.6 & 94.6 & 73.6 & 93.9 & 96.5 & 95.5 & 73.7 & 90.3 & 82.8 & 95.4 & 97.7 & 95.9 & 98.6 & 77.6 & 88.7 & 78.0 & 98.3 & 89.0 & 90.6\\
HCP~\cite{hcp} & 98.6 & 97.1 & 98.0 & 95.6 & 75.3 & \textbf{94.7} & 95.8 & 97.3 & 73.1 & 90.2 & 80.0 & 97.3 & 96.1 & 94.9 & 96.3 & 78.3 & 94.7 & 76.2 & 97.9 & 91.5 & 90.9\\
RNN-Attention~\cite{rnn_attention} & 98.6 & 97.4 & 96.3 & 96.2 & 75.2 & 92.4 & 96.5 & 97.1 & 76.5 & 92.0 & 87.7 & 96.8 & 97.5 & 93.8 & 98.5 & 81.6 & 93.7 & 82.8 & 98.6 & 89.3 & 91.9\\
Atten-Reinforce~\cite{reinforce} & 98.6 & 97.1 & 97.1 & 95.5 & 75.6 & 92.8 & 96.8 & 97.3 & 78.3 & 92.2 & 87.6 & 96.9 & 96.5 & 93.6 & 98.5 & 81.6 & 93.1 & 83.2 & 98.5 & 89.3 & 92.0\\
\hline
\hline
VGG (Binary) & 98.3 & 97.1 & 96.1 & 96.7 & 75.0 & 91.4 & 95.8 & 95.4 & 76.7 & 92.1 & 85.1 & 96.7 & 96.0 & 95.3& 97.8 & 77.4 & 93.1 & 79.7 & 97.9 & 89.3 & 91.1\\
VGG (Re-weighted) & 99.4 & 97.4 & 98.0 & 97.0 & 77.9 & 92.4 & 96.8 & 97.8 & 80.8 & 93.4 & 87.2 & 98.0 & 97.3 & 95.8 & 98.8 & 79.4 & 95.3 & 82.2 & 99.1 & 91.4 & 92.8\\
\hline
ML-GCN (Binary) & 99.6 & 98.3 & 97.9 & 97.6 & 78.2 & 92.3 & 97.4 & 97.4 & 79.2 & 94.4 & 86.5 & 97.4 & 97.9 & 97.1 & 98.7 & 84.6 & 95.3 & 83.0 & 98.6 & 90.4 & 93.1\\
ML-GCN (Re-weighted) & 99.5 & \textbf{98.5} & \textbf{98.6} & \textbf{98.1} & 80.8 & 94.6 & \textbf{97.2} & 98.2 & \textbf{82.3} & \textbf{95.7} & 86.4 & 98.2 & \textbf{98.4} & \textbf{96.7} & \textbf{99.0} & \textbf{84.7} & \textbf{96.7} & \textbf{84.3} & 98.9 & \textbf{93.7} & \textbf{94.0}\\
\hline
\end{tabular}
\label{table:voc}
\end{table*}

\paragraph{Results on VOC 2007}

PASCAL Visual Object Classes Challenge (VOC 2007)~\cite{voc} is another popular dataset for multi-label recognition. It contains 9,963 images from 20 object categories, which is divided into train, val and test sets. Following~\cite{reinforce,rnn_attention}, we use the \emph{trainval} set to train our model, and evaluate the recognition performance on the {test} set. In order to compare with other state-of-the-art methods, we report the results of average precision (AP) and mean average precision (mAP). 

The results of VOC 2007 are presented in Table~\ref{table:voc}. Because the results of many previous works on VOC 2007 are based on the VGG model~\cite{verydeep}. For fair comparisons, we also report the results using VGG models as the base model. It is apparent to see that, our proposed method observes improvements upon the previous methods. Concretely, the proposed ML-GCN with our re-weighted scheme obtains $94.0\%$ mAP, which outperforms state-of-the-art by $2\%$. Even using VGG model as the base model, we can still achieve better results ($+0.8\%$). Also, consistent with the results on MS-COCO, the re-weighed scheme enjoys better performance than the binary correlation matrix on VOC as well.


\subsubsection{Ablation Studies}

In this section, we perform ablation studies from four different aspects, including the sensitivity of ML-GCN to different types of word embeddings, effects of $\tau$ in correlation matrix binarization, effects of $p$ for correlation matrix re-weighting, and the depths of GCN.

\paragraph{ML-GCN under different types of word embeddings}

\begin{figure}[t]
	\centering
	\includegraphics[width=0.95\columnwidth]{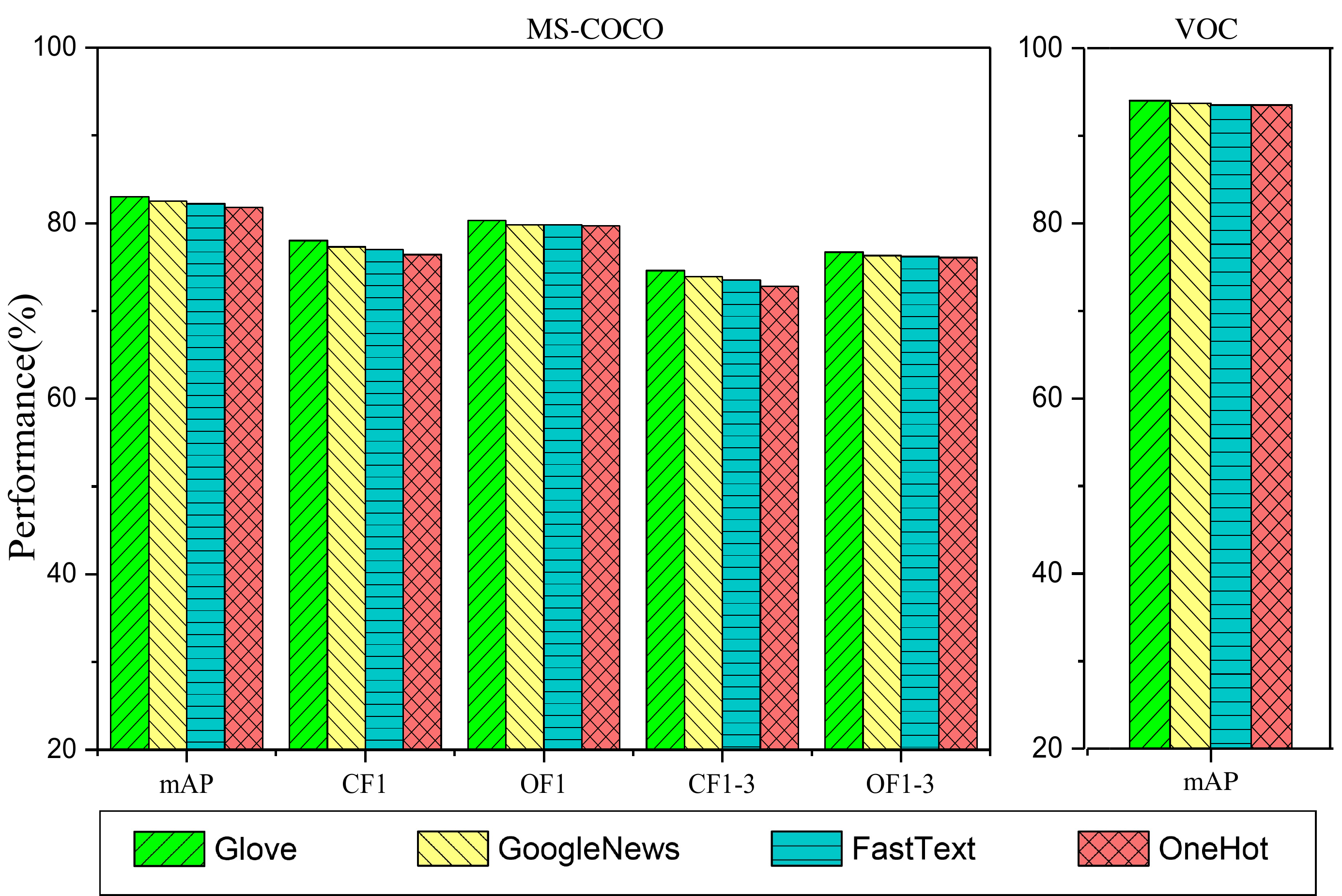}
	\caption{Effects of different word embedding approaches. It is clear to see that, different word embeddings will hardly affect the accuracy, which reveals our improvements do not absolutely come from the semantic meanings derived from word embeddings, rather than our ML-GCN.}
	\label{fig:word}
\end{figure}

By default, we use Glove~\cite{glove} as label representations, which serves as the inputs of the stacked GCNs for learning the object classifiers. In this part, we evaluate the performance of ML-GCN under other types popular word representations. Specifically, we investigate four different word embedding methods, including GloVe~\cite{glove}, GoogleNews~\cite{googlenews}, FastText~\cite{fasttext} and the simple one-hot word embedding. Fig.~\ref{fig:word} shows the results using different word embeddings on MS-COCO and VOC 2007. As shown, we can see that when using different word embeddings as GCN's inputs, the multi-label recognition accuracy will not be affected significantly. In addition, the observations (especially the results of one-hot) justify that the accuracy improvements achieved by our method do not absolutely come from the semantic meanings derived from word embeddings. Furthermore, using powerful word embeddings could lead to better performance. One possible reason may be that the word embeddings~\cite{glove,googlenews,fasttext} learned from large text corpus maintain some semantic topology. That is, for semantic related concepts, their embeddings are close in the embedding space. Our model can employ these implicit dependencies, and further benefit multi-label image recognition. 


\paragraph{Effects of different threshold values $\tau$}

\begin{figure}[t]
	\centering
	\subfloat[Comparisons on MS-COCO.] {\includegraphics[width=0.5\columnwidth]{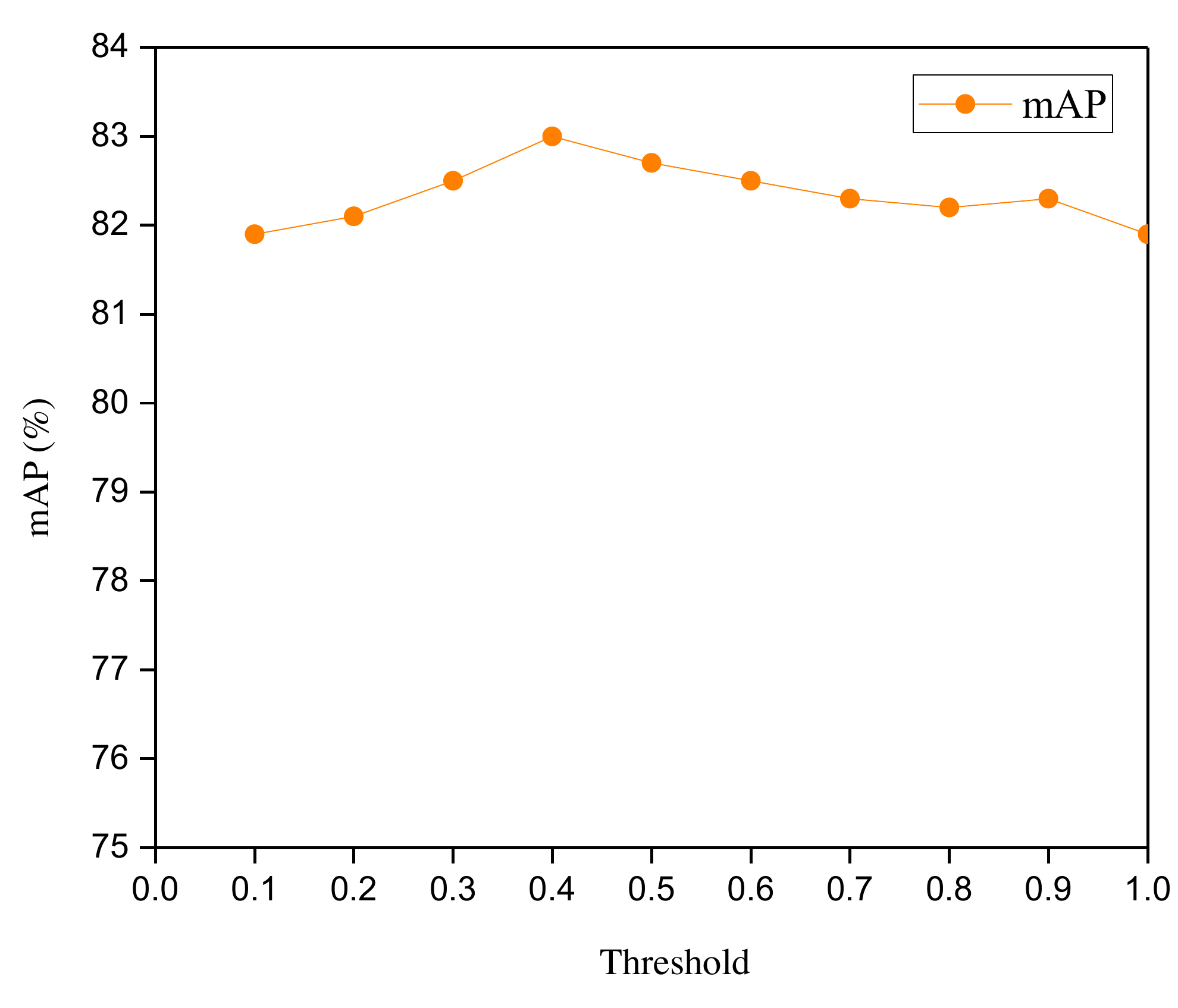} }
	\subfloat[Comparisons on VOC 2007.]  {\includegraphics[width=0.5\columnwidth]{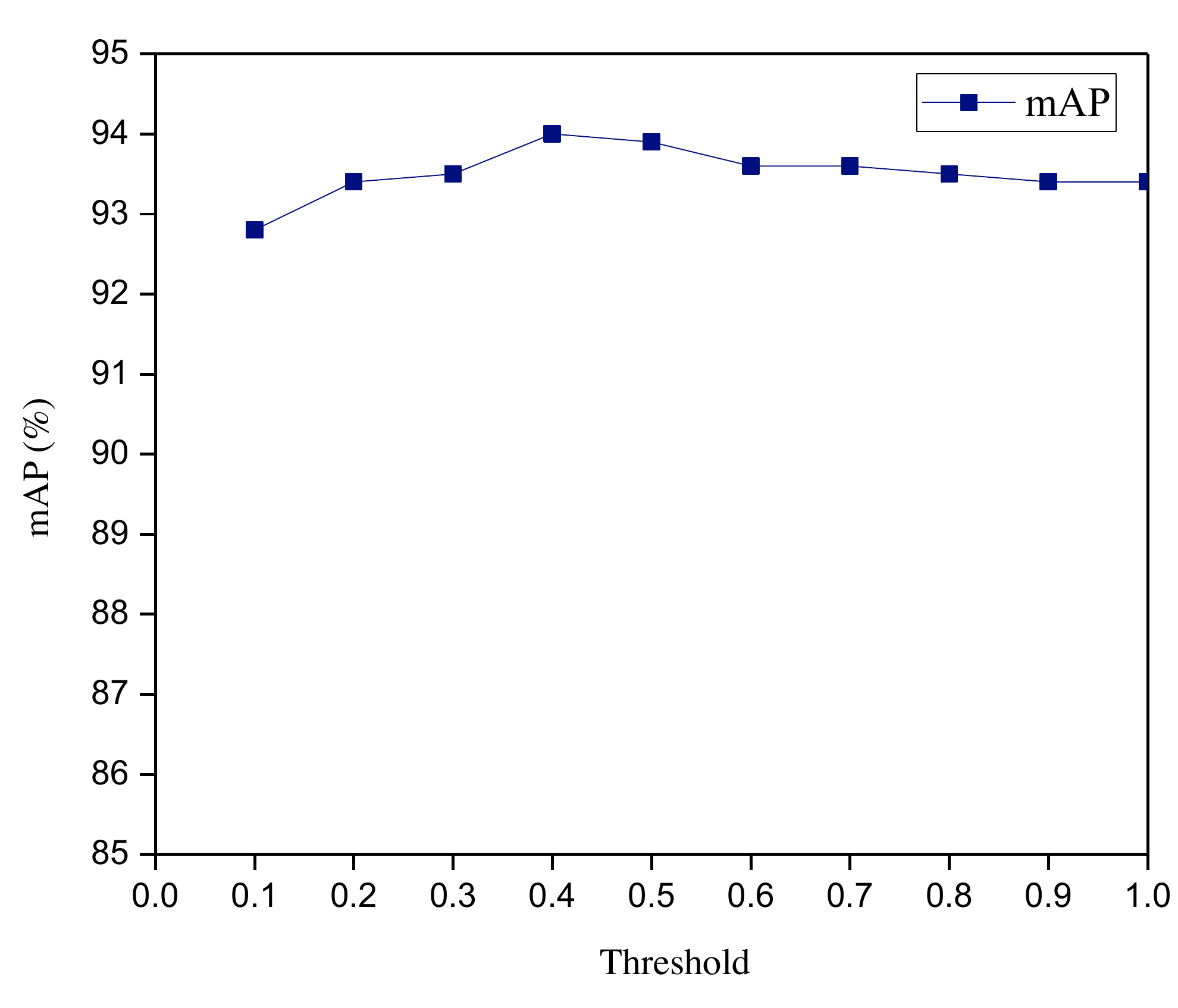} }
	\caption{Accuracy comparisons with different values of $\tau$.}
	\label{fig:threshold}
\end{figure}

We vary the values of the threshold $\tau$ in Eq.~(\ref{eq:binary}) for correlation matrix binarization, and show the results in Fig.~\ref{fig:threshold}. Note that, if we do not filter any edges, the model will not converge. Thus, there is no result for $\tau = 0$ in that figure. As shown, when filtering out the edges of small probabilities (\emph{i.e.}, noisy edges), the multi-label recognition accuracy is boosted. However, when too many edges are filtered out, the accuracy drops since correlated neighbors will be ignored as well. The optimal value of $\tau$ is $0.4$ for both MS-COCO and VOC 2007.

\paragraph{Effects of different $p$ for correlation matrix re-weighting}

\begin{figure}
	\centering
	\subfloat[Comparisons on MS-COCO.]  {\includegraphics[width=0.5\columnwidth]{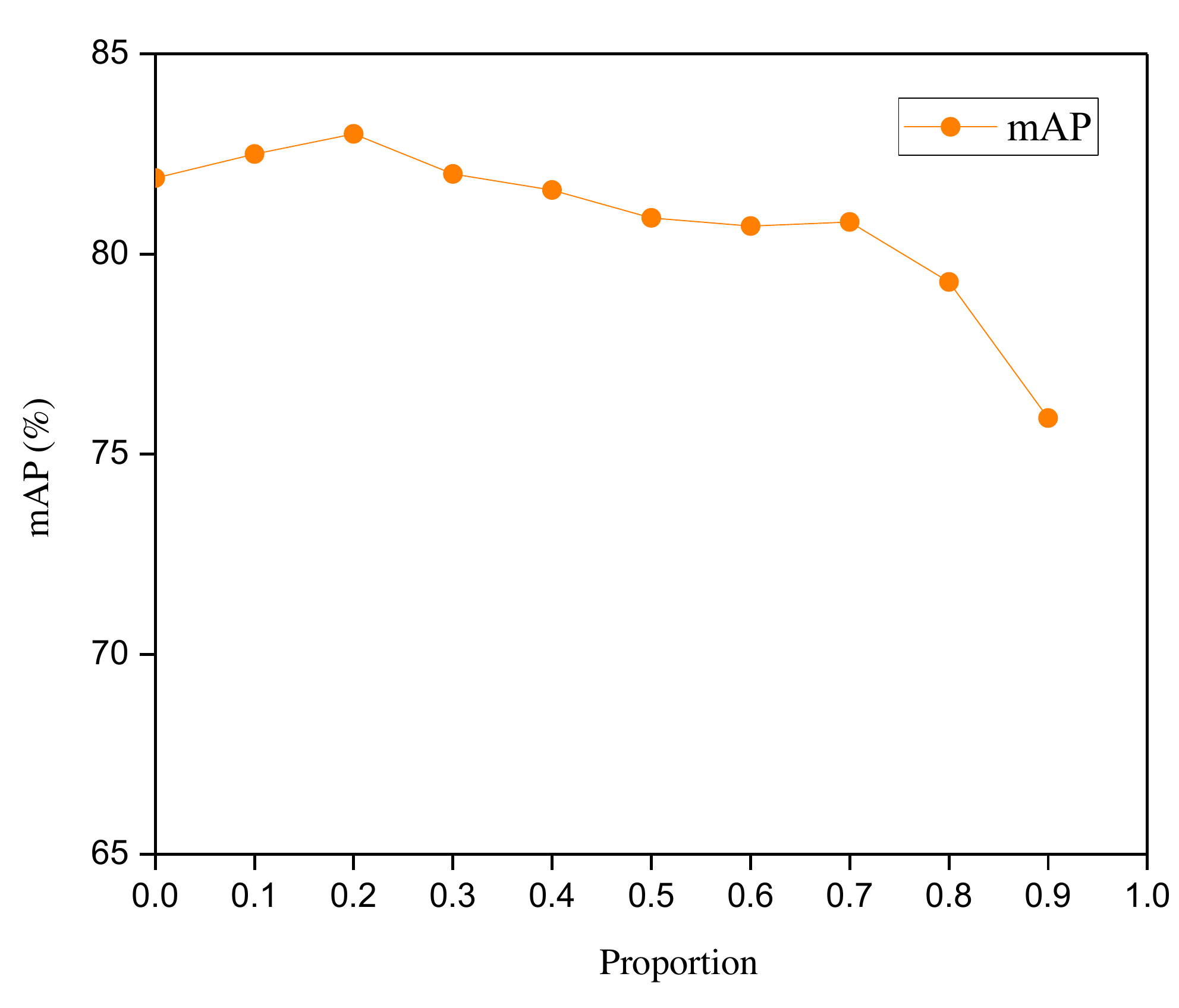} }
	\subfloat[Comparisons on VOC 2007.]  {\includegraphics[width=0.5\columnwidth]{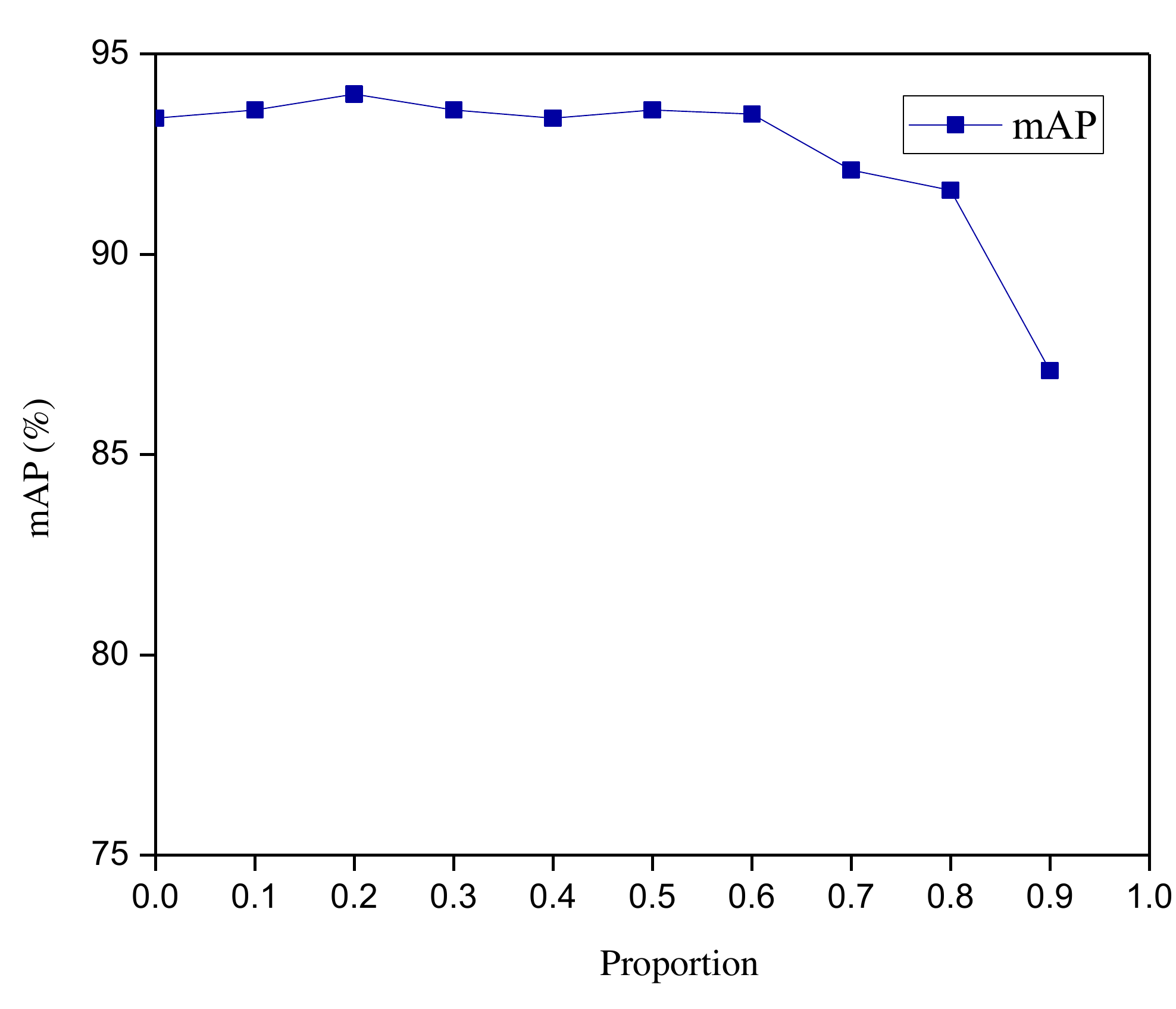} }
	\caption{Accuracy comparisons with different values of $p$. Note that, when $p=1$, the model does not converge.}
	\label{fig:proportion}
\end{figure}

To explore the effects of different values of $p$ in Eq.~(\ref{eq:reweight}) on multi-label classification accuracy, we change the values of $p$ in a set of $\left\{0,0.1, 0.2, \ldots, 0.9, 1\right\}$, as depicted in Fig.~\ref{fig:proportion}. Generally, this figure shows the importance of balancing the weights between a node itself and the neighborhood when updating the node feature in GCN. In experiments, we choose the optimal value of $p$ by cross-validations. We can see that when $p=0.2$, it can achieve the best performance on both MS-COCO and VOC 2007. If $p$ is too small, nodes (labels) of the graph can not get sufficient information from correlated nodes (labels). While, if $p$ is too large, it will lead to over-smoothing. 

Another interesting observation is that, when $p=0$, we can obtain mAPs of $81.67\%$ on MS-COCO and $93.15\%$ on VOC 2007, which still outperforms existing methods. Note that when $p=0$, we essentially do not explicitly incorporate the label correlations. The improvement is benefited from that our ML-GCN model learns the object classifiers from the prior label representations through a shared GCN based mapping function, which implicitly models label dependencies as discussed in Sec.~\ref{sec:motivation}

\begin{figure*}[t!]
	\centering
	\includegraphics[width=0.85\textwidth]{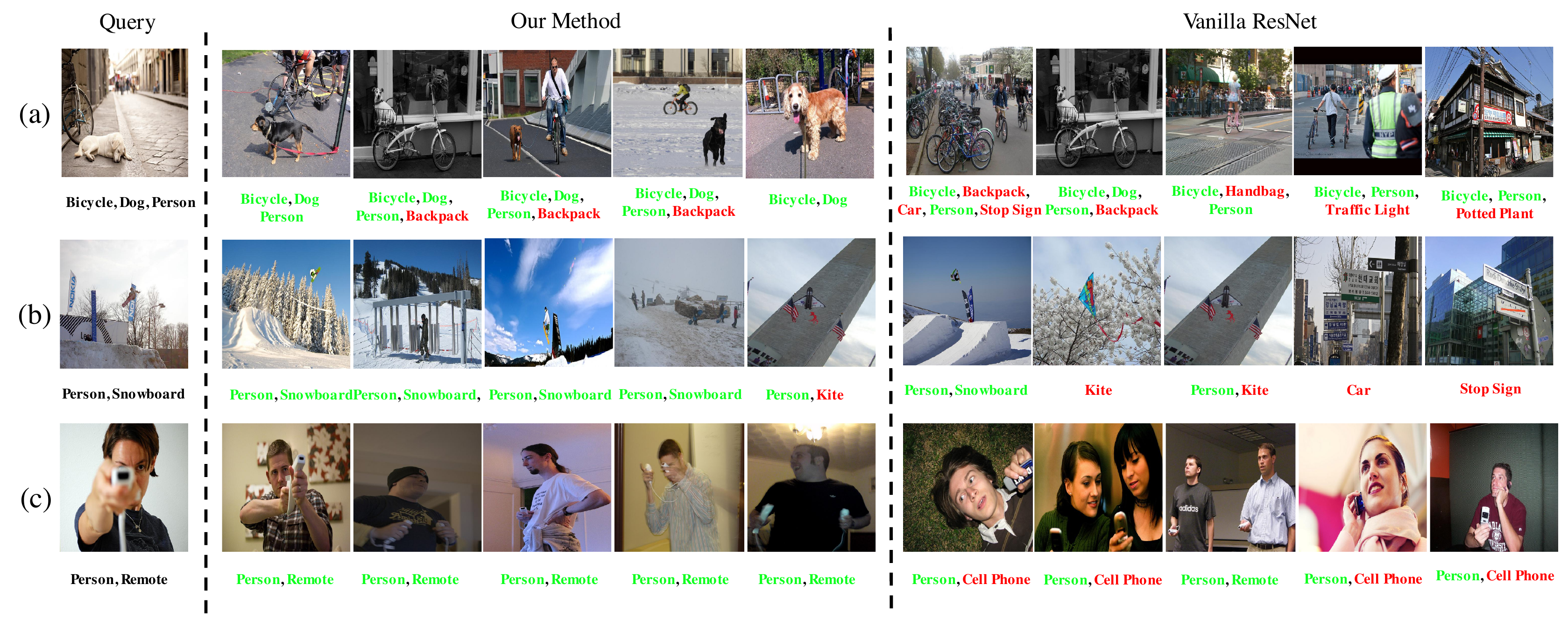}
	\vspace{-0.5em}
	\caption{Top-5 returned images with the query image. The returned results on the left are based on our proposed ML-GCN, while the results on the right are vanilla ResNet. All results are sorted in the ascending order according to the distance from the query image.}
	\label{fig:knn}
\end{figure*}

\paragraph{The deeper, the better?}

We show the performance results with different numbers of GCN layers for our model in Table~\ref{table:depth}. For the three-layer model, the output dimensionalities are $1024$, $1024$ and $2048$ for the sequential layers, respectively. For the four-layer model, the dimensionalities are $1024$, $1024$, $1024$ and $2048$. As shown, when the number of graph convolution layers increases, multi-label recognition performance drops on both datasets. The possible reason for the performance drop may be that when using more GCN layers, the propagation between nodes will be accumulated, which can result in over-smoothing.


\begin{table}[t]
\footnotesize
\centering
\caption{Comparisons with different depths of GCN in our model.}
\vspace{0.1cm}
\begin{tabular}{|c||c|c|c||c|c|c|c|}
\cline{1-6} \cline{8-8} 
        & \multicolumn{5}{c|}{MS-COCO} && VOC \\
\cline{1-6} \cline{8-8}
\multirow{2}{*}{$\sharp$ Layer} & \multicolumn{3}{c||}{{All}} & \multicolumn{2}{c|}{{Top-3}} && All \\
\cline{2-6} \cline{8-8} & mAP & CF1 & OF1 & CF1 & OF1 && mAP \\
\cline{1-6} \cline{8-8}
2-layer & \textbf{83.0} & \textbf{78.0} & \textbf{80.3} & \textbf{74.6} & \textbf{76.7} && \textbf{94.0}\\
3-layer & 82.1 & 76.9 & 79.7 & 73.7 & 76.2 && 93.6\\
4-layer & 81.1 & 76.4 & 79.4 & 72.5 & 75.8 && 93.0\\
\cline{1-6} \cline{8-8}
\end{tabular}
\label{table:depth}
\end{table}

\subsection{Classifier Visualization}

The effectiveness of our approach has been quantitatively evaluated through comparisons to existing methods and detailed ablation studies. In this section, we visualize the learned inter-dependent classifiers to show if meaningful semantic topology can be maintained. 



\begin{figure}[h!]
	\centering
	\subfloat[t-SNE on the learned inter-dependent classifiers by our model.]  {\includegraphics[width=0.85\columnwidth]{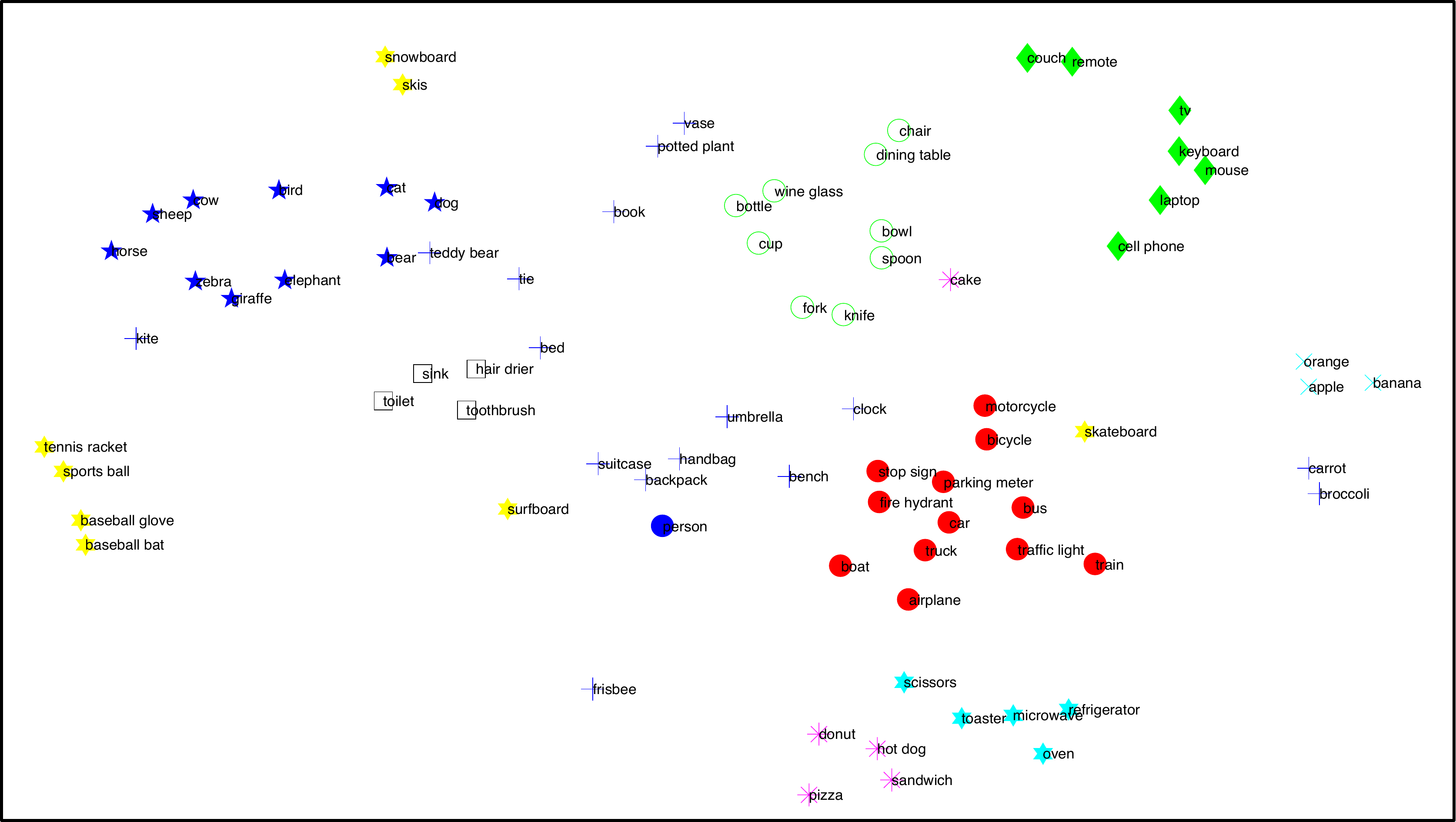} }\\
	\subfloat[t-SNE on the classifiers by the vanilla ResNet.]  {\includegraphics[width=0.85\columnwidth]{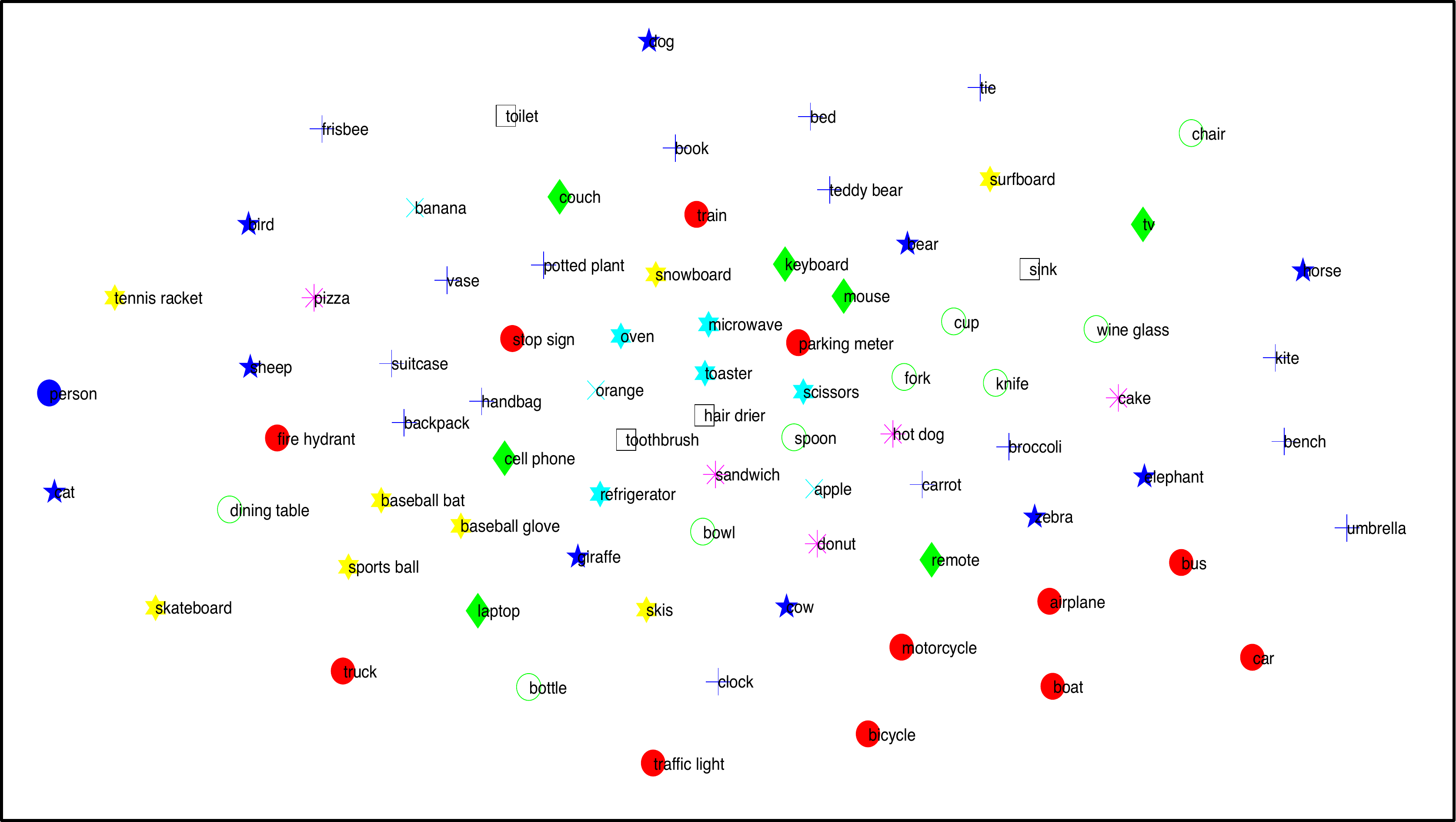} }\\
	\includegraphics[width=0.86\columnwidth]{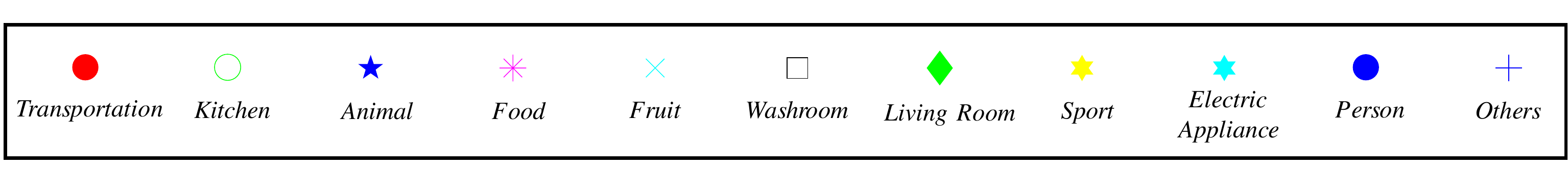}
	\caption{Visualization of the learned inter-dependent classifiers by our model and vanillia classifiers of ResNet on MS-COCO.}
	\vspace{-0.5em}
	\label{fig:classifier}
	\vspace{0cm}
\end{figure}

In Fig.~\ref{fig:classifier}, we adopt the t-SNE~\cite{tsen} to visualize the classifiers learned by our proposed ML-GCN, as well the classifiers learned through vanilla ResNet (\ie, parameters of the last fully-connected layer). It is clear to see that, the classifiers learned by our method maintain meaningful semantic topology. Specifically, the learned classifiers exhibit cluster patterns. Classifiers (of ``\texttt{car}'' and ``\texttt{truck}'') within one super concept (``\texttt{transportation}''), tend to be close in the classifier space. This is consistent with common sense, which indicates that the classifiers learned by our approach may not be limited to the dataset where the classifiers are learned, but may enjoy generalization capacities. On the contrary, the classifiers learned through vanilla ResNet uniformly distribute in the space and do not shown any meaningful topology. This visualization further shows the effectiveness of our approach in modeling label dependencies.


\subsection{Performance on image retrieval}

Apart from analyzing the learned classifiers, we further evaluate if our model can learn better image representations. We conduct an image retrieval experiment to verify this. Specifically, we use the $k$-NN algorithm to perform content-based image retrieval to validate the discriminative ability of image representations learned by our model. Still, we choose the features from vanilla ResNet as the baseline. We show the top-5 images returned by $k$-NN. The retrieval results are presented in Fig.~\ref{fig:knn}. For each query image, the corresponding returned images are sorted in the ascending order according to the distance to the query image. We can clearly observe that our retrieval results are obviously better than the vanilla ResNet baseline. For example,  in Fig.~\ref{fig:knn} (c), the labels of the images returned by our approach almost exactly match the labels of the query image. It can demonstrate that our ML-GCN can not only effectively capture label dependencies to learn better classifiers, but can benefit image representation learning as well in multi-label recognition.

\section{Conclusion}

Capturing label dependencies is one crucial issue for multi-label image recognition. In order to model and explore this important information, we proposed a GCN based model to learn inter-dependent object classifiers from prior label representations, \eg, word embeddings. To explicitly model the label dependencies, we designed a novel re-weighted scheme to construct the correlation matrix for GCN by balancing the weights between a node and its neighborhood for node feature update. This scheme can effectively alleviate over-fitting and over-smoothing, which are two key factors hampering the performance of GCN. Both quantitative and qualitative results validated the advantages of our ML-GCN.


{\small
\bibliographystyle{ieee}
\bibliography{egbib}
}

\end{document}